\documentclass[10pt,twocolumn,letterpaper]{article}

\usepackage{iccv}      % To produce the REVIEW version
\usepackage[accsupp]{axessibility}  % Improves PDF readability for those with disabilities.

\usepackage{amsmath}
\usepackage{booktabs}
\usepackage{tikz}
\usepackage{multirow}

\usepackage{makecell}
\usepackage{arydshln}
\usepackage{comment}
\usepackage{xcolor}
\usepackage{soul}
\usetikzlibrary{positioning, arrows.meta, shapes.multipart}

\definecolor{iccvblue}{rgb}{0.21,0.49,0.74}
\usepackage[pagebackref,breaklinks,colorlinks,allcolors=iccvblue]{hyperref}

\def\paperID{VQualA-23} % *** Enter the Paper ID here
\def\confName{ICCV}
\def\confYear{2025}

\title{A Lightweight Ensemble-Based Face Image Quality Assessment Method with Correlation-Aware Loss}

\author{
MohammadAli Hamidi$^{1}$,
Hadi Amirpour$^{2}$,
Luigi Atzori$^{1}$,
Christian Timmerer$^{2}$ \\
$^{1}$DIEE, University of Cagliari, CNIT, University of Cagliari, 09123 Cagliari, Italy \\
$^{2}$ Department of Information Technology (ITEC), Alpen-Adria Universität Klagenfurt, Austria \\
{\tt\small mohammadali.hamidi@unica.it, hadi.amirpour@aau.at, l.atzori@unica.it, christian.timmerer@aau.at}
}

\begin{document}
\maketitle

\begin{abstract}
Face image quality assessment (FIQA) plays a critical role in face recognition and verification systems, especially in uncontrolled, real-world environments. Although several methods have been proposed, general-purpose no-reference image quality assessment techniques often fail to capture face-specific degradations. Meanwhile, state-of-the-art FIQA models tend to be computationally intensive, limiting their practical applicability. We propose a lightweight and efficient method for FIQA, designed for the perceptual evaluation of face images in the wild. Our approach integrates an ensemble of two compact convolutional neural networks, MobileNetV3-Small and ShuffleNetV2, with prediction-level fusion via simple averaging. To enhance alignment with human perceptual judgments, we employ a correlation-aware loss (MSECorrLoss), combining mean squared error (MSE) with a Pearson correlation regularizer. Our method achieves a strong balance between accuracy and computational cost, making it suitable for real-world deployment. Experiments on the VQualA FIQA benchmark demonstrate that our model achieves a Spearman rank correlation coefficient (SRCC) of 0.9829 and a Pearson linear correlation coefficient (PLCC) of 0.9894, remaining within competition efficiency constraints.
\end{abstract} 


\section{Introduction}
\label{sec:intro}

Face photographs are among the most frequently captured, shared, and processed images today. Constant streams from smartphone cameras and social-media filters to e-gate checkpoints, payment apps, and city-wide CCTV, place facial data at the center of tasks such as rapid identity checks, access control, cosmetic retouching, and online interaction~\cite{yang_face_2019,hong_unsupervised_2020,jawahar_scale-varying_2019}. The economic and social stakes are substantial: modern face recognition secures airports, unlocks phones, organizes photo libraries, and powers targeted advertising, while police and forensic units rely on clear faces to trace identity and responsibility. The accuracy and, therefore, the reliability of these human and machine-driven tasks rest on the visual and algorithmic quality of each image. Evaluating whether a face photo is ``fit for purpose” has therefore become almost as critical as recognizing the face itself.

FIQA is the gatekeeper that decides whether a portrait is suitable for automated face-recognition pipelines~\cite{schlett_face_2022}. In biometric systems, blurred, poorly lit, or occluded faces are screened out before matching, which prevents sharp drops in recognition accuracy and keeps false matches in check. Because recognition performance in the wild is highly sensitive to pose, illumination, and occlusion, production pipelines now rely on FIQA to filter low-grade captures that would otherwise overwhelm even state-of-the-art (SOTA) networks~\cite{zhang_study_2023,cuellar_accuracy_2025}. Beyond security, many visual-media workflows, such as face restoration, deep-fake detection, and cosmetic retouching analysis, also need a face-specific quality score; generic image metrics either disagree with human perception or scale poorly~\cite{rathgeb_facial_2022}. By providing an automatic, face-aware, and task-relevant judgment, FIQA protects downstream algorithms, reduces evaluation costs, and ultimately strengthens user trust in any application that depends on reliable facial imagery~\cite{fu_deep_2022}.

%FIQA condenses every face image into a single numeric score that represents its utility to automated face-recognition systems, not its aesthetic appeal~\cite{best-rowden_learning_2018}. Because this score is optimized for machine performance, a crisp, well-lit profile portrait can receive a low rating, while a noisy but perfectly frontal shot may score high--highlighting a gap between algorithmic usefulness and what humans consider “high quality”~\cite{terhorst_face_2020}. 
FIQA is essential not only for ensuring reliable performance in automated face recognition systems but also for maintaining perceptual quality from a human perspective~\cite{best-rowden_learning_2018,terhorst_face_2020}. Conventional no-reference image-quality metrics, designed to model human perception, correlate only weakly with recognition success~\cite{mittal_no-reference_2012}, whereas modern face-specific FIQA methods track recognition accuracy far more closely~\cite{terhorst_ser-fiq_2020}. Closing the remaining gap is the aim of perceptual FIQA: metrics that respect the operational needs of recognition pipelines while aligning with the sharpness, naturalness, and dignity cues people intuitively associate with a good portrait.

General no-reference image-quality (IQA) metrics~\cite{zhou_perceptual_2025} such as BRISQUE~\cite{mittal_blindreferenceless_2011}, NIQE~\cite{mittal_making_2013}, and PIQE~\cite{venkatanath_n_blind_2015} were designed to mirror human judgments of sharpness or naturalness, yet repeated benchmarking shows that their scores correlate only weakly with face-recognition accuracy under real-world variations in pose, illumination, and occlusion~\cite{fu_deep_2022}. This shortcoming has sparked a shift toward perceptual~\cite{wei_recent_2024} FIQA, which estimates a portrait’s recognition utility directly. Early models learned quality from match scores, unsupervised ones from embedding stability; newer methods encode it inside identity vectors, and attention or diffusion-based variants sharpen key regions for real-time, accuracy-aligned scores.

Despite their impressive accuracy, most SOTA perceptual FIQA models depend on heavyweight backbones, repeated forward passes, or iterative diffusion loops, driving inference into the multi-GFLOP range and pushing latency beyond real-time budgets, raising an obstacle for mobile or embedded deployment. To remove this bottleneck, we present a lightweight, efficiency-oriented FIQA approach that predicts perceptual quality with a compact network footprint, making it suitable for resource-constrained environments while retaining the performance benefits of modern perceptual scoring. 

Our key contributions are summarized as follows:
\textbf{First}, we design a compact two-branch ensemble that fuses MobileNetV3-Small and ShuffleNetV2 predictions, leveraging their complementary representational strengths of different architectures to achieve high accuracy at a sub-million-parameter scale. \textbf{Second}, we introduce MSECorrLoss, a correlation-aware objective that simultaneously minimizes MSE and maximizes Pearson correlation with human ratings, thereby tightening alignment with perceptual ground truth. \textbf{Third}, we integrate an effective Test-Time Augmentation (TTA) strategy that averages predictions across multiple augmented views of each input image, leading to improved robustness and stability in quality assessment predictions.

%-------------------------------------------------------------------------
\section{Related works}
Full-reference metrics such as PSNR, SSIM~\cite{wang_image_2004}, VIF~\cite{sheikh_image_2006}, and LPIPS~\cite{zhang_unreasonable_2018} provide reliable estimates when a pristine ground-truth frame is available, because they quantify error visibility or perceptual similarity relative to that reference. In practice, however, the original image is often either unavailable or compromised, particularly in the case of historical mugshots, body-worn-camera footage, and user-generated social media videos seldom include an undistorted counterpart, making it impossible to apply full-reference measures. Under these conditions, no-reference (blind) quality assessment becomes indispensable for both real-time monitoring and offline curation of facial data streams. 

FIQA exhibits the same two-track structure. In most operational settings, such as surveillance video, body-worn cameras, selfie unlock, and social media uploads, the clean reference portrait is either missing or already degraded, forcing practitioners to rely on no-reference metrics. Generic blind scores such as BRISQUE~\cite{mittal_blindreferenceless_2011}, NIQE~\cite{mittal_making_2013}, PIQE~\cite{venkatanath_n_blind_2015}, and NIMA~\cite{talebi_nima_2018} provide a quick first pass, while face-specific FIQA methods refine the judgment by explicitly modelling pose, illumination, and expression. 

SOTA FIQA methods predominantly employ deep neural networks to estimate face image quality, leveraging diverse architectures and supervision sources. CR‑FIQA~\cite{boutros_cr-fiqa_2023} learns a regression head on top of embeddings of iResNet backbones (50/100), trained to predict a relative classifiability score derived from a sample’s angular proximity to its class center and nearest negative class center. This method demonstrates superior correlation performance across eight benchmarks compared to prior SOTA. CLIB‑FIQA~\cite{ou_clib-fiqa_2024}  enhances the anchoring-based training approach by integrating vision-language alignment via CLIP embeddings and calibrating label confidence using joint distributions of objective quality factors (blur, pose, expression, occlusion, illumination). This joint learning and calibration framework corrects unreliable quality anchors from recognition models and significantly improves performance on eight datasets. 

IFQA~\cite{ifqa}, which uses an adversarial restoration framework combined with a per-pixel discriminator to generate interpretable spatial quality maps highlighting critical facial regions (eyes, nose, mouth), demonstrating strong alignment with human perception and improving downstream training when used as an objective function. Going beyond face-specific systems, TOPIQ~\cite{topiq} introduces a cognitive-inspired, coarse-to-fine transformer architecture that uses semantic cues to guide attention toward local distortion regions. %Though not tailored to faces, TOPIQ achieves notable gains in both reference and no-reference settings with only 13\% of FLOPs compared to heavyweight models, suggesting that semantic-awareness could be similarly valuable for FIQA. 

\begin{figure*}[t]
	\centering
	\includegraphics[width=6.7in]{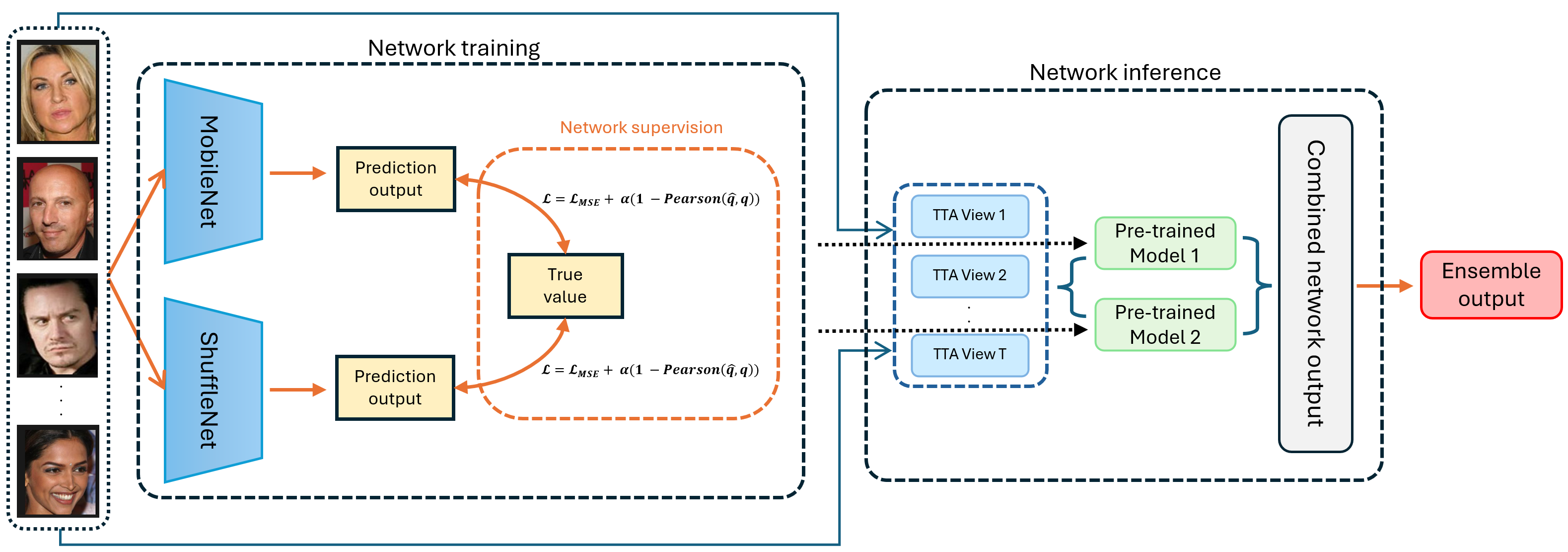}
    \caption{Overview of the proposed ensemble-based FIQA architecture. During training (left), two lightweight CNNs learn to predict face image quality scores using a correlation-aware loss function that combines MSE and Pearson correlation. During inference (right), each input image undergoes Test-Time Augmentation (TTA), producing \(T\) augmented views. Both models process each augmentation, and predictions are averaged first across augmentations, then across models, yielding the final quality score.
}
    \label{fig: pros-arc}
\end{figure*}

Despite major strides~\cite{schlett_face_2022}, many SOTA methods rely on heavy architectures that pose challenges for low-latency or on-device deployment. This observation motivates our work: a lightweight, no-reference FIQA approach that preserves strong correlation with human perceptual judgments while meeting real-world efficiency constraints. 

\section{Proposed Method}
We propose an ensemble-based FIQA approach that combines the predictions of two lightweight and efficient convolutional neural networks (CNNs), namely MobileNetV3-Small and ShuffleNetV2, both fine-tuned for the FIQA task. A simple averaging strategy is employed to aggregate the outputs of the two models (as illustrated in Figure~\ref{fig: pros-arc}). These compact models extract meaningful features from face images, generating high-dimensional representations that effectively capture perceptual quality information. Notably, the total number of trainable parameters across both networks is around 2 million, making our approach highly efficient. 
This dual-level averaging scheme mitigates overfitting and captures diverse perceptual cues, improving robustness to variations in input images.

During training, the models' predictions are supervised using the correlation-aware loss, which combines Mean Squared Error (MSE) with a Pearson correlation loss. This encourages not only accurate predictions but also alignment with the relative ranking of perceptual quality scores, which is crucial in perceptual quality tasks.

\begin{figure}[t]
	\centering
	\includegraphics[width=0.4\textwidth]{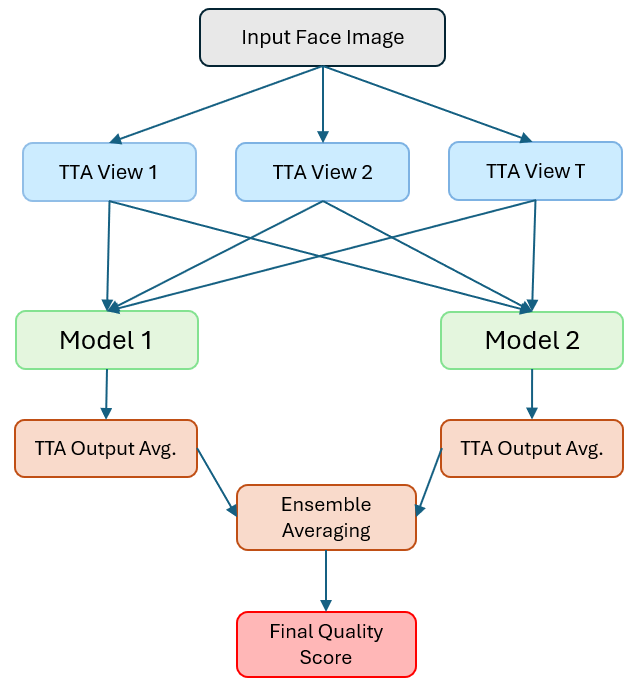}
    \caption{TTA process. The input face image is augmented into multiple views. Each model processes the views, followed by per-model TTA averaging, and final ensemble fusion.}
    \label{fig: tta_aug}
\end{figure}

At inference time, to enhance robustness, we apply the Test-Time Augmentation (TTA) approach \cite{TTA}, where multiple augmented views of each image are created and processed through the ensemble. The final quality score is obtained by aggregating predictions across both augmentations and models, as shown in Figure \ref{fig: tta_aug}. 

\begin{figure*}[t]
	\centering
	\includegraphics[width=6.5in]{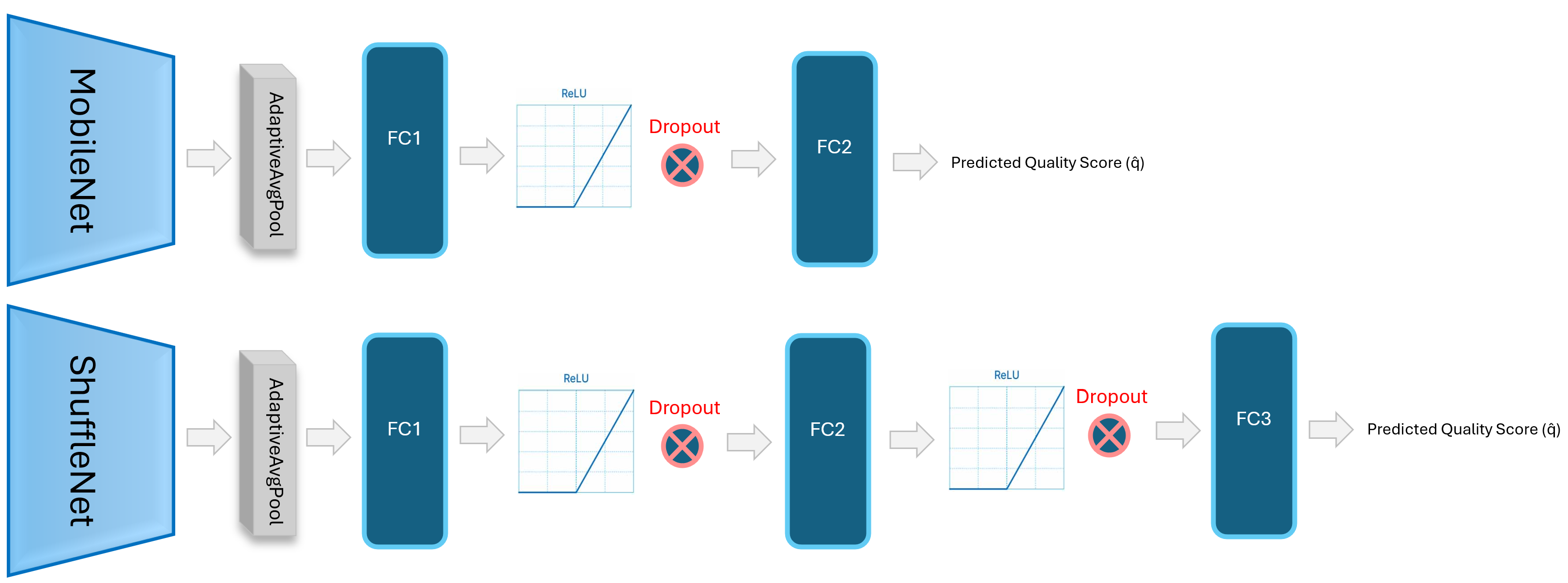}
    \caption{Architectures of the MobileNet and ShuffleNet networks used for the FIQA task. Each network consists of a pre-trained backbone followed by an adaptive average pooling layer and an MLP module. The MLP module includes one or more linear layers, ReLU activations, and dropouts with rates of 0.2, culminating in a scalar output representing the predicted quality score. These configurations were designed to capture discriminative features while maintaining computational efficiency.}
    \label{fig: finetuned-arc}
\end{figure*}

\subsection{Architecture}
Given an input image \(I\), each model \(m \in \{1, ..., M\} \) in the ensemble predicts a quality score $\hat{q}_m(I)$. The final quality prediction is obtained by:

\begin{equation}
\hat{q}(I) = \frac{1}{M} \sum_{m=1}^M \frac{1}{T} \sum_{t=1}^T \hat{q}_{m,t}(I)
\end{equation}

where $\hat{q}_{m,t}(I)$ is the prediction of model $m$ on the $t^{th}$ augmented version of image \(I\). Here, \(M=2\) is the number of models in the ensemble, and \(T\) is the number of augmented versions \( \mathcal{T} =\{T_1(I), T_2(I),..., T_T(I)\}\), which is defined to 3 in our case. Each model processes these transformed versions, and the final prediction is obtained by averaging across both augmentations and models.

% In the training phase, input images are rescaled to $600 \times 416$ without any extra augmentation. We also tried multiple augmentation methods, and no obvious impact was observed.

Due to the high performance and low inference cost, two lightweight CNN networks were utilized for extracting features from the face images, including MobileNetV3-Small and ShuffleNetV2, pre-trained on ImageNet and fine-tuned for the FIQA task by replacing their final classification layer with a Multi-Layer Perceptron (MLP) tailored to output a single perceptual quality score  (shown in Figure~\ref{fig: finetuned-arc}).

Let $I$ be the input face image. And $\mathbf{f}=GAP(Backbone(I)) \in \mathbb{R}^{576}$ be the global average pooled feature vector from the MobileNet backbone. The final quality prediction $\hat{q} \in \mathbb{R}$ is computed as follows:

% \begin{equation}
% \hat{q} = \mathbf{W}_2 \cdot \text{Dropout} \left( \text{ReLU} \left( \mathbf{W}_1 \cdot \text{GAP} \left( \text{Backbone}(I) \right) + b_1 \right) \right) + b_2
% \end{equation}

% \begin{equation}
% \hat{q} = W_2 \alpha (ReLU (W_1  (\text{GAP} (\text{Backbone}(I))) +b_1 ) + b_2
% \end{equation}

\begin{equation}
\begin{aligned}
\mathbf{h}_1 &= \text{Dropout}\left( \text{ReLU}\left( \mathbf{W}_1 \mathbf{f} + \mathbf{b}_1 \right) \right) \\
\hat{q} &= \mathbf{W}_2 \mathbf{h}_1 + b_2
\end{aligned}
\end{equation}

where,
\( \quad \mathbf{W}_1 \in \mathbb{R}^{576 \times 288} \) and \( \quad b_1 \in \mathbb{R}^{288} \) are trainable parameters of the first linear layer FC1, and  \( W_2 \in \mathbb{R}^{288 \times 1} \) and \( \quad b_2 \in \mathbb{R}^{1} \) are trainable parameters of the second linear layer FC2.

Similarly, let $\mathbf{f} \in \mathbb{R}^{1024}$ be the global average pooled output of the ShuffleNet backbone. The final prediction $\hat{q} \in \mathbb{R}$ is computed via:

% \begin{equation}
% \hat{q} = \mathbf{W}_3 \cdot \text{Dropout} \left( \text{ReLU} \left( \mathbf{W}_2 \cdot \text{Dropout} \left( \text{ReLU} \left( \mathbf{W}_1 \cdot \text{GAP}(\text{Backbone}(I)) + b_1 \right) \right) + b_2 \right) \right) + b_3
% \end{equation}

\begin{equation}
\begin{aligned}
\mathbf{h}_1 &= \text{Dropout}\left( \text{ReLU}\left( \mathbf{W}_1 \mathbf{f} + \mathbf{b}_1 \right) \right) \\
\mathbf{h}_2 &= \text{Dropout}\left( \text{ReLU}\left( \mathbf{W}_2 \mathbf{h}_1 + \mathbf{b}_2 \right) \right) \\
\hat{q} &= \mathbf{W}_3 \mathbf{h}_2 + b_3
\end{aligned}
\end{equation}

where,
\( \quad \mathbf{W}_1 \in \mathbb{R}^{1024 \times 512} \) and \( \quad b_1 \in \mathbb{R}^{512} \) are trainable parameters of the first linear layer FC1, \( W_2 \in \mathbb{R}^{512 \times 256} \) and \( \quad b_2 \in \mathbb{R}^{256} \) are trainable parameters of the second linear layer FC2, and \( W_3 \in \mathbb{R}^{256 \times 1} \) and \( \quad b_2 \in \mathbb{R}^{1} \) are trainable parameters of the third linear layer FC3.

The models are trained to minimize a joint loss that combines mean squared error (MSE) with a correlation-based term:

\begin{equation}
\mathcal{L} = \mathcal{L}_{MSE}(q_i, \hat{q}_i) + \alpha \mathcal{L}_{Corr}(q_i, \hat{q}_i)
\end{equation}

where $\hat{q}_i$ and $q_i$ denote the ground-truth and predicted scores for the $i-th$ training sample, respectively, and $\alpha$ is a hyperparameter that balances the two terms.
 
The MSE component is defined as:
\begin{equation}
    \mathcal{L}_{\text{MSE}} = \frac{1}{N} \sum_{i=1}^N \left( q_i - \hat{q}_i \right)^2
\end{equation} 

and, the correlation loss is defined as:
\begin{equation}
    \mathcal{L}_{\text{Corr}} = 1 - \text{Pearson}(q_i, \hat{q}_i)
\end{equation}

where the Pearson correlation is given by:
\begin{equation}
\text{Pearson}(\hat{q}, q) = 
\frac{
\sum_{i=1}^{N} \left( \hat{q}_i - \bar{\hat{q}} \right) \left( q_i - \bar{q} \right)
}{
\sqrt{ \sum_{i=1}^{N} \left( \hat{q}_i - \bar{\hat{q}} \right)^2 } 
\sqrt{ \sum_{i=1}^{N} \left( q_i - \bar{q} \right)^2 }
}
\end{equation}

Here, \(\bar{\hat{q}}\) and \(\bar{q}\) represent the mean of the predicted and ground-truth scores, respectively.

\begin{comment}

During inference, the input face image \(I\) undergoes test-time augmentation, producing \(T\) transformed versions \( \mathcal{T} =\{T_1(I), T_2(I),..., T_T(I)\}\). Each model processes these versions, and the final prediction is obtained by averaging across augmentations and across models: 

% Let’s consider the given input image \(I\), the prediction is computed as:

\begin{equation}
\hat{q}(I) = \frac{1}{M} \sum_{m=1}^M \frac{1}{T} \sum_{t=1}^T \hat{q}_{m,t}(I)
\end{equation}

where $\hat{q}_{m,t}(I)$ is the prediction of model $m$ on the $t^{th}$ augmented version of image \(I\). Here, \(M=2\) is the number of models in the ensemble, and \(T\) is the number of augmentations which is defined to 3 in our case.
\end{comment}

\begin{table*}
\centering
\caption{Performance comparison with the state-of-the-art approaches on the VQualA FIQA challenge dataset.}
\label{tab:results}
\resizebox{0.73\textwidth}{!}{%
\centering
\begin{tabular}{|c|c|c|c|c|}
\hline 
{\textbf{Type}} & {\textbf{Approaches}} & \textbf{SRCC} & \textbf{PLCC} & \textbf{Final Score}  \\ \cline{1-5}
\multirow{7}{*}{\makecell[c]{\textbf{General}\\\textbf{NR}}} & NIMA \cite{talebi_nima_2018} & 0.5839 & 0.7649 & 0.6744 \\ \cline{2-5}
 & DB-CNN \cite{db-cnn}& 0.5324 & 0.7833 & 0.6578  \\ \cline{2-5}
 & QualiCLIP \cite{qualiclip}& 0.5324 & 0.7833 & 0.6578 \\ \cline{2-5}
 & PIQE \cite{venkatanath_n_blind_2015} & 0.6090 & 0.8122 & 0.7106 \\ \cline{2-5}
 & NIQE \cite{mittal_making_2013} & 0.6914 & 0.8574 & 0.7744 \\ \cline{2-5}
 & BRISQUE \cite{mittal_blindreferenceless_2011} & 0.6465 & 0.8149 & 0.7307 \\ \cline{2-5}
 & MANIQA \cite{yang2022maniqa} & 0.7790 & 0.8918 & 0.8354 \\ \cline{2-5}

\hdashline \multirow{4}{*}{\makecell[c]{\textbf{Face}\\\textbf{NR}}} 
 & TOPIQ\_Face \cite{topiq} & 0.8623 & 0.9266 & 0.8945 \\ \cline{2-5}
 & TOPIQ\_Swin\_Face \cite{topiq} & 0.9156 & 0.9416 & 0.9286 \\ \cline{2-5}
 & IFQA \cite{ifqa} & 0.3962 & 0.4258 & 0.4110 \\ \cline{2-5} 
 & \textbf{Ours} & \textbf{0.9829} & \textbf{0.9894} & \textbf{0.9862} \\ \cline{2-5}
\hline
\end{tabular}
}
\end{table*}

\section{Experimental Results}
To investigate the performance of the proposed FIQA method, we have conducted experiments on the VQualA FIQA competition dataset \footnote{https://codalab.lisn.upsaclay.fr/competitions/23017} and compared the achieved performance with that of SOTA methods.

\subsection{Dataset}
The VQualA FIQA challenge dataset is a large-scale, in-the-wild collection of facial images annotated with subjective quality scores. It comprises approximately 30,000 images for training, 1,000 for validation, and 1,000 for testing. Due to the competition rule, no one had access to the test set; instead, the validation set is utilized for our experiments in this work. The dataset reflects real-world variability in facial appearance and acquisition conditions, offering a diverse range of quality levels. Importantly, the image sizes are not normalized, with widths ranging from 200 to 1,000 pixels, which adds to the complexity and realism of the quality assessment task.

\subsection{Implementation details}
The proposed approach was implemented on a workstation featuring an Intel Xeon  418H CPU, an NVIDIA RTX 6000 Ada Gen. GPU, and 512 GB of RAM. The model was developed using the PyTorch deep learning framework. The input images were resized to \(600\times416\) without applying additional data augmentation. Although various augmentation strategies were tested, they did not yield any significant improvement in performance. The initial learning rate of the models was set to $5 \times 10^{-4}$ with a weight decay of $1 \times 10^{-4}$ in every 5 epochs of training. We observed that applying a lower learning rate to the backbones facilitated more stable optimization and faster convergence using the ADAM optimizer \cite{Adam}. The batch size was set to 64, and training followed a maximum of 30 epochs for each model. The final models contain approximately 2 million parameters and 0.4985 GFLOPs per sample, keeping our model lightweight and highly efficient.

To enhance robustness in the inference, we applied data augmentation techniques, including Random Horizontal Flip (p = 1.0) and Random Vertical Flip(p = 1.0), to introduce data variations during the inference.
The datasets were divided with a $80\%/20\%$ splitting rate for the training and validation sets, respectively.

\subsection{Comparing with the State-of-the-art}
%\MOHAMMAD{We should describe how we trained SOTA models.}
%\MOHAMMAD{What about this: "We evaluated SOTA models using their available pretrained weights without retraining, to benchmark their generalization ability on our dataset".}

To assess the effectiveness of our proposed method, we compare its performance against several SOTA no-Reference image quality assessment approaches, including both general-purpose and face-specific models. For SOTA models, pretrained weights were used without any additional retraining.

The evaluation is conducted on the VQualA FIQA Challenge dataset, using standard correlation metrics (SRCC, PLCC) and their average as the final score defined by the challenge organizers~\cite{ma2025vquala}. As shown in Table~\ref{tab:results}, our method significantly outperforms existing baselines across all metrics, demonstrating the benefit of our ensemble learning strategy, correlation-based loss function, and TTA process tailored for face image quality estimation. In particular, a relevant performance with a large margin improvement is observed over the top-performing face-specific approach (TOPIQ Swin Face) with higher SRCC (0.9829 vs. 0.9156) and PLCC (0.9894 vs. 0.9416), indicating a 0.058 surpassing in the final averaged score (0.9862 vs. 0.9286). Moreover, our approach inherits the practical application with approximately 2 million parameters and 0.4985 GFLOPs in resource-constrained environments. These results validate the framework as a high-capacity yet efficient model that achieves state-of-the-art performance while maintaining a favorable computational cost.

\begin{table*}[h]
\caption{Ablation study on the impact of model architecture, loss function, and TTA strategy.}
\label{tab:ablation}
\centering
\resizebox{1\linewidth}{!}{%
\begin{tabular}{|l|c|c|c|c|c|c|}
\hline
\textbf{Variant} & \textbf{Model(s)} & \textbf{Loss} & \textbf{Other} & \textbf{SRCC} & \textbf{PLCC} & \textbf{Final Score} \\
\hline
Baseline A & MobileNet & MSE & – & 0.9662 & 0.9773 & 0.9718 \\
Baseline B & ShuffleNet & MSE & – & 0.9638 & 0.9726 & 0.9682 \\
+ Ensemble & MobileNet + ShuffleNet & MSE & – & 0.9747 & 0.9836 & 0.9792 \\
+ Corr-Aware Loss & MobileNet + ShuffleNet & MSECorrLoss & – & 0.9774 & 0.9868 & 0.9821 \\
+ TTA & MobileNet + ShuffleNet & MSECorrLoss & TTA & \textbf{0.9829} & \textbf{0.9894} & \textbf{0.9862} \\
\hline
\end{tabular}
}
\end{table*}

\subsection{Ablation study}
To thoroughly assess the contribution of each component in our proposed model, we conducted a series of ablation studies. These experiments isolate the impact of individual design choices, including the ensemble method, loss functions, and TTA strategy. By systematically modifying or removing these components, we demonstrate how each element contributes to the overall performance improvement. The following results in Table \ref{tab:ablation} are reported using SRCC, PLCC, and their average as the final score, aligning with the evaluation protocol of the VQualA FIQA Challenge.

\subsubsection{Impact of ensemble learning method}
To evaluate the effectiveness of the ensemble learning strategy in our proposed framework, we conducted an ablation study comparing the performance of individual models with their ensemble. The architecture leverages two compact yet complementary backbones, MobileNet and ShuffleNet, each trained and fine-tuned independently on the FIQA task.

To improve robustness and generalization, we constructed a deep ensemble model by averaging the predictions of both fine-tuned backbones. The ensemble inference combines model outputs at the decision level, leveraging complementary learning patterns of the two networks. This reflects consistent improvement over the individual models, particularly in SRCC, demonstrating better rank correlation with human judgments. The improvement gained through ensemble validation tests the hypothesis that combining multiple lightweight models, even with limited capacity, can outperform each individually. The ensemble helps reduce prediction variance, compensates for weaknesses in each model, and leads to a more stable and accurate prediction of perceptual quality.

% \begin{table}
% \centering
% \caption{Ablation study on the effect of ensemble learning.}
% \label{tab:ensemble_ablation}
% \resizebox{0.8\columnwidth}{!}{%
% \begin{tabular}{lccc}
% \toprule
% \textbf{Model} & \textbf{SRCC} & \textbf{PLCC} & \textbf{Final Score} \\
% \midrule
% MobileNetV3 & 0.9665 & 0.9775 & 0.9720 \\
% ShuffleNet & 0.9640 & 0.9730 & 0.9685 \\
% Ensemble & \textbf{0.9740} & \textbf{0.9840} & \textbf{0.9790} \\
% \bottomrule
% \end{tabular}
% }
% \end{table}

\subsubsection{Impact of Loss Function}
To evaluate the effect of our proposed loss function on performance, we compared two alternatives: (i) the standard MSE loss and (ii) our combined MSECorrLoss, which integrates the MSE term with a Pearson correlation-based regularizer. This formulation explicitly encourages the model not only to minimize prediction error but also to better preserve the monotonic relationship between predicted and ground truth MOS values, which is a key property in quality assessment tasks.

As shown in Table \ref{tab:ablation}, training with MSECorrLoss consistently led to improved performance over using MSELoss alone. Specifically, SRCC improved from 0.9747 to 0.9774, and PLCC increased from 0.9836 to 0.9868. These gains result in a higher overall final score (mean of SRCC and PLCC), demonstrating the effectiveness of the proposed loss in better aligning the model’s output with subjective human ratings.

% \begin{table}[!t]
% \centering
% \caption{Ablation study on the effect of loss functions.}
% \label{tab:loss_ablation}
% \resizebox{0.8\columnwidth}{!}{%
% \begin{tabular}{lccc}
% \toprule
% \textbf{Loss Function} & \textbf{SRCC} & \textbf{PLCC} & \textbf{Final Score} \\
% \midrule
% MSELoss  & 0.9721 & 0.9823 & 0.9772 \\
% MSECorrLoss (Ours)   & \textbf{0.9755} & \textbf{0.9854} & \textbf{0.9804} \\
% \bottomrule
% \end{tabular}
% }
% \end{table}

\subsubsection{Impact of Test-Time Augmentation (TTA)}
TTA enhances model robustness by averaging predictions over multiple augmented versions of the input. In our setup, TTA includes horizontal flips and slight color variations during inference. As shown in Table~\ref{tab:ablation}, enabling TTA yields improved correlation metrics by reducing prediction variance and mitigating overfitting to specific input patterns. The performance gain is particularly noticeable in SRCC, demonstrating better rank consistency across samples.

% \begin{table}[!t]
% \centering
% \caption{Ablation study on the effect of Test-Time Augmentation (TTA).}
% \label{tab:tta_ablation}
% \resizebox{0.8\columnwidth}{!}{%
% \begin{tabular}{lccc}
% \toprule
% \textbf{TTA Strategy} & \textbf{SRCC} & \textbf{PLCC} & \textbf{Final Score} \\
% \midrule
% Without TTA  & 0.9730 & 0.9832 & 0.9781 \\
% With TTA     & \textbf{0.9756} & \textbf{0.9855} & \textbf{0.9806} \\
% \bottomrule
% \end{tabular}
% }
% \end{table}

\section{Conclusion}
In this work, we proposed a lightweight ensemble-based framework for Face Image Quality Assessment (FIQA) combining the strengths of two efficient backbones, including MobileNet and ShuffleNet, with a novel correlation-aware loss function. Through a series of ablation studies, we demonstrated the effectiveness of each module, including the impact of ensemble learning, the proposed MSECorrLoss function, and the TTA strategy. Our method achieves a SOTA final score of 0.9862 on the VQualA FIQA Challenge dataset, significantly outperforming existing NR-IQA models. These results highlight the potential of simple yet well-orchestrated architectural and training strategies in advancing robust, accurate, and computationally efficient FIQA systems. Future work may explore modality-specific calibration and adaptation to in-the-wild face quality scenarios.

\section*{Acknowledgments}
This work has been partially supported by the European Union - Next Generation EU under the Italian National Recovery and Resilience Plan (NRRP), Mission 4, Component 2, Investment 1.3, CUP C29J24000300004, partnership on “Telecommunications of the Future” (PE00000001 - program “RESTART”), by the European Union under the Italian NRRP of NextGenerationEU, "Sustainable Mobility Center" Centro Nazionale per la Mobilità Sostenibile, CNMS, CN\_00000023), and by Italian NRRP - M4C1 - Inv. 3.4 and M4C1 - Inv. 4.1, Ministerial Decree no. 351/2022.
This work has also been partially funded by the European Union (SPIRIT, 101070672). Views and opinions expressed are however those of the author(s) only and do not necessarily reflect those of the European Union. Neither the European Union nor the granting authority can be held responsible for them. The SPIRIT project has received funding from the Swiss State Secretariat for Education, Research and Innovation (SERI).

\small
\bibliographystyle{ieeenat_fullname}
\bibliography{main,references}
\end{document}